\documentclass[conference]{IEEEtran}
\IEEEoverridecommandlockouts
\usepackage{cite}
\usepackage{amsmath,amssymb,amsfonts}
\usepackage{algorithmic}
\usepackage{graphicx}
\usepackage{textcomp}

\usepackage{url}
\usepackage{booktabs}
\usepackage{ulem}
\usepackage{tabularx}
\usepackage[table]{xcolor}
\usepackage{colortbl}
\usepackage{multirow}
\def\BibTeX{{\rm B\kern-.05em{\sc i\kern-.025em b}\kern-.08em
    T\kern-.1667em\lower.7ex\hbox{E}\kern-.125emX}}
\begin{document}

\title{CoF: Coarse to Fine-Grained Image Understanding for Multi-modal Large Language Models\\
\thanks{* contribute equally to this work and $\dagger$ 
 corresponding authors.}
}

\author{

\IEEEauthorblockN{Yeyuan Wang$^{*,1}$, Dehong Gao$^{*,2}$, Bin Li$^1$, Rujiao Long$^3$, Lei Yi$^3$, \\ Xiaoyan Cai$^{\dagger,1}$, Libin Yang$^{\dagger,2}$, Jinxia Zhang$^{4,5}$, Shanqing Yu$^{6,7}$, Qi Xuan$^{6,7}$} \\
\IEEEauthorblockA{
\textit{$^1$School of Automation, Northwestern Polytechnical University, Xi’an, Shaanxi, China}\\
\textit{$^2$School of Cybersecurity, Northwestern Polytechnical University, Xi'an, Shaanxi, China}\\
\textit{$^3$Alibaba Group, Hangzhou, Zhejiang, China} \\
\textit{$^4$The Key Laboratory of Measurement and Control of CSE, Ministry of Education,} \\
\textit{School of Automation, Southeast University, Nanjing 210096, China} \\
\textit{$^5$Advanced Ocean Institute of Southeast University, Nantong 226010, China} \\
\textit{$^6$Zhejiang University of Technology, Hangzhou, Zhejiang, China} \\
\textit{$^7$Binjiang Institute of Artificial Intelligence, Hangzhou, Zhejiang, China} \\
\textit{\{wangyeyuan, libin0225\}}@mail.nwpu.edu.cn, 
\textit{\{dehong.gdh, xiaoyanc, libiny\}}@nwpu.edu.cn \\
\textit{\{rujiao.lrj, yilei.yi\}}@alibaba-inc.com,
\textit{jinxiazhang}@seu.edu.cn,
\textit{\{yushanqing, qixuan\}}@zjut.edu.cn
} 
} 

\maketitle

\begin{abstract}
The impressive performance of Large Language Model (LLM) has prompted researchers to develop Multi-modal LLM (MLLM), which has shown great potential for various multi-modal tasks. 
However, current MLLM often struggles to effectively address fine-grained multi-modal challenges. 
We argue that this limitation is closely linked to the models' visual grounding capabilities. 
The restricted spatial awareness and perceptual acuity of visual encoders frequently lead to interference from irrelevant background information in images, causing the models to overlook subtle but crucial details. 
As a result, achieving fine-grained regional visual comprehension becomes difficult. 
In this paper, we break down multi-modal understanding into two stages, from Coarse to Fine (CoF). 
In the first stage, we prompt the MLLM to locate the approximate area of the answer. 
In the second stage, we further enhance the model's focus on relevant areas within the image through visual prompt engineering, adjusting attention weights of pertinent regions. 
This, in turn, improves both visual grounding and overall performance in downstream tasks. 
Our experiments show that this approach significantly boosts the performance of baseline models, demonstrating notable generalization and effectiveness. 
Our CoF approach is available online at \url{https://github.com/Gavin001201/CoF}.
\end{abstract}


\begin{figure*}[!htbp]
\centerline{\includegraphics[width=\linewidth]{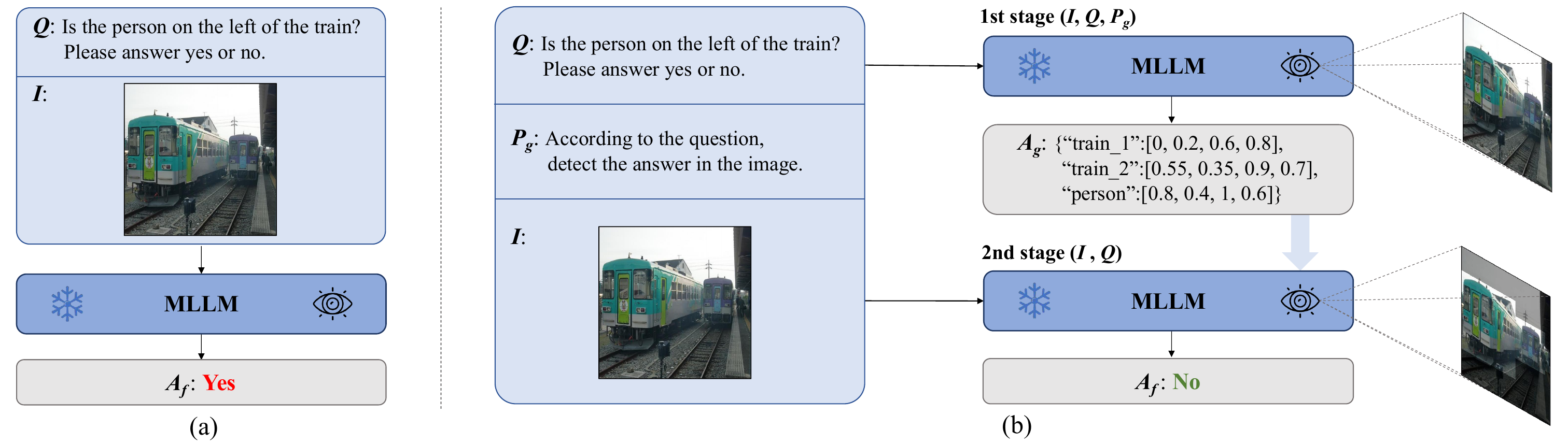}}
\caption{Overview of (a) baseline approach and (b) our CoF approach.
CoF consists of two stages: (1) Location Grounding and (2) Attention Reweighting.
In the first stage, the MLLM takes the question \textit{Q}, the grounding prompt \textit{P$_\text{g}$}, and image \textit{I} as input and obtain the coarse-grained coordinates of the answer region.
In the second stage, we reweight the attention score of the answer region according to the coarse-grained coordinates obtained in the first stage to guide the model to focus on the answer region.}
\label{framework}
\end{figure*}

\section{Introduction}
Large Language Model (LLM) has demonstrated remarkable capabilities in natural language processing (NLP)~\cite{touvron2023llama, chiang2023vicuna}. 
Researchers have further advanced these models by integrating them with visual perception modules, resulting in Multi-modal Large Language Model (MLLM)~\cite{zhu2023minigpt, liu2024improved, dai2023instructblip}. 
Experimental evidence indicates that MLLM performs effectively across a variety of multi-modal tasks~\cite{ge2023makingllamadrawseed, team2023gemini, achiam2023gpt}. 
However, the limited perceptual abilities of current visual perception models hinder their capacity for fine-grained understanding~\cite{fashionbert, wei2023varyscalingvisionvocabulary, yuan2024osprey, wang2024enhancing}.

Prominent open-source MLLM typically utilizes the model pretrained with contrastive learning (e.g., CLIP~\cite{clip}) as visual perception component~\cite{liu2024improved}.
However, it is reported that this instance-level contrast~\cite{clip, align} does not work well in fine-grained perception~\cite{zhuge2021kaleidobert, filip, singh2023coarsetofine}.
This limitation is transferred to MLLM that integrate these visual perception models, rendering them more susceptible to distractions from irrelevant information~\cite{chen2023position,wei2023varyscalingvisionvocabulary}.
Consequently, they frequently struggle to identify subtle yet critical target information, negatively affecting their performance in fine-grained downstream tasks and increasing the likelihood of generating ``hallucinations".

Considerable research has attempted to address this limitation~\cite{xuan2024pink, jain2024vcoder}. 
These studies focus on enhancing the visual perception ability of MLLM by increasing image resolution or introducing auxiliary information~\cite{yuan2024osprey, xuan2024pink, jain2024vcoder}. 
Moreover, researchers have observed that the CLIP visual encoder tends to prioritize content within specific marked areas~\cite{red_circle, yang2024fine, cai2024vip}.
Building on this observation, they have employed visual prompt engineering to guide the model's attention towards specific areas, optimizing its capture of target information~\cite{yang2023set}.
Subsequent efforts have integrated these findings into MLLM, specifically by creating region-level image-text pair datasets with distinct visual markers and adjusting model parameters to enhance MLLM fine-grained understanding capabilities~\cite{yuan2024osprey, guo2024regiongpt, zhang2024prompt}.
However, these methods often lead to higher computational and require substantial human labor. 
Thus, there is an urgent need for a more efficient approach.

To comprehend complex visual information, humans typically focus on specific regions or details within a given sample.
For the situation shown in Fig.~\ref{framework}, when asked with providing a detailed description of a local area or inquiring about the attributes of a small target object, humans generally scan the entire image, locate the approximate position of the target, and then concentrate on it.
In contrast, most MLLM process image information at a fixed granularity (e.g., Fig.~\ref{framework}(a))~\cite{alayrac2022flamingo, moon2023anymal, wu2023next}, making it challenging to capture subtle but critical information that may be difficult to detect visually.  
To mimic efficient human-like reasoning, models need to identify the target image region containing essential visual details and pay more attention to it to capture fine-grained critical visual information (e.g., Fig.~\ref{framework}(b)). 
This is difficult for current MLLM to deal with, causing them to be easily affected by background noise or trying to find answers from text prompts and amplifying the influence of language priors, resulting in poor performance on fine-grained tasks and hallucination problems.

In this paper, we propose a Coarse-to-Fine-grained (CoF) multi-modal understanding approach aimed at improving the fine-grained perception capabilities of MLLM.
Different from current Chain-of-Thought (CoT) approaches that directly guide the input and output of the model, our approach operates in the latent variable space.
The proposed CoF approach consists of two main stages.
The first stage involves identifying the target area(s) within the given image, while the second stage focuses on comprehending its(their) semantic information. 
We simplify fine-grained multi-modal understanding into two manageable stages by breaking down the image understanding process. 
The initial identification of the target area allows for greater concentration on this critical region in the subsequent stage, thereby reducing interference from irrelevant information. 
This refined focus increases the likelihood of achieving an accurate understanding of the target area.
In our specific implementation, we first prompt the MLLM to ascertain the coordinates of the target area, subsequently assigning higher attention weights to the visual tokens in this region to precisely direct the model's focus. 
This method significantly improves its fine-grained understanding capabilities. 
Extensive experiments validate the effectiveness of our approach.

The primary contributions of this paper are as follows: 
(1) We introduce the CoF (Coarse-to-Fine) approach, which streamlines image understanding into two stages: coarse-grained localization and fine-grained perception.
(2) We integrate visual prompt engineering technology to bolster the fine-grained understanding capabilities of MLLM in a low-resource manner.
(3) We conduct comprehensive experiments that demonstrate CoF's robust fine-grained understanding capabilities across a wide range of downstream tasks.

\section{METHOD}

\begin{figure*}[htbp]
\centerline{  \includegraphics[width=\linewidth]{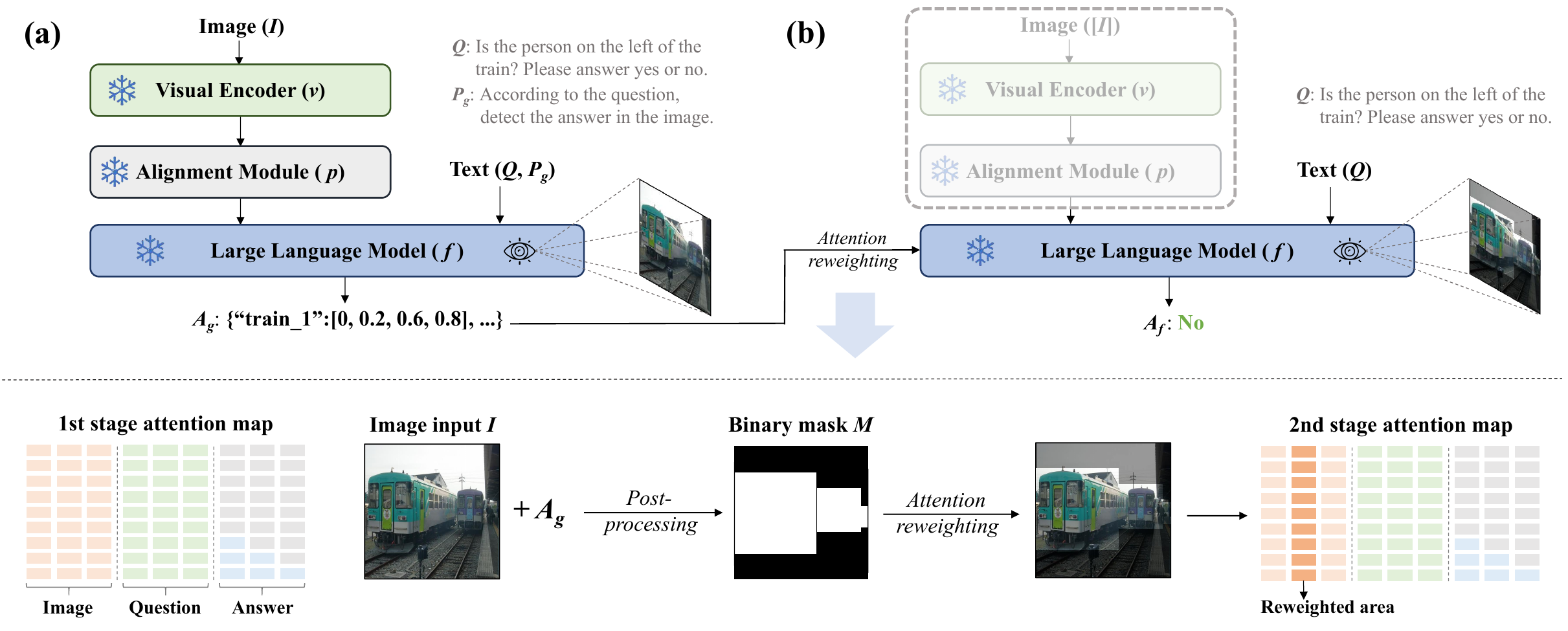}}
\caption{Overview of CoF approach.
(a) In the first stage, the model determines the answer area in the input image based on the question. 
(b) Then the output coordinates are post-processed and converted into a binary mask matrix, and the attention map of the input image is reweighted according to the mask matrix.}
\label{method}
\end{figure*}

As illustrated in Fig.~\ref{method}, our CoF comprises two primary stages: (1) the coarse-grained location grounding stage and (2) the fine-grained attention reweighting stage. 
This section delves into a comprehensive description of the approach.

\subsection{Preliminaries}
MLLM typically consists of a visual encoder $v_{\phi}(\cdot)$, a LLM decoder $f_{\theta}(\cdot)$, and an image-text alignment module $p_{\epsilon}(\cdot)$ (parameterized by $\phi$, $\theta$, $\epsilon$ respectively).
Given an image \textit{I} and a textual prompt \textit{P}, the image \textit{I} is first encoded into visual embeddings via $v_{\phi}(\cdot)$, which are then mapped to a shared LLM embedding space through $p_{\epsilon}(\cdot)$ and result in a set of visual tokens $e_{v}$. 
The LLM decodes the output response \textit{R} in a sequence manner, which is formulated as:
\begin{equation}
R = f_{\theta}(p_{\epsilon}(v_{\phi}(\textit{I})), 
\textit{P})
\end{equation}

For decoder, the Transformer-based LLM is centered on the attention module and model the relationship between $e_{v}$ and $e_{t}$ through the attention mechanism.
Specifically, the model computes the attention map $A$ to represent the relationship between $e_{v}$ and $e_{t}$ using the following formula:
\begin{equation}
A = softmax(\frac{[e_{v}, e_{t}] \cdot ([e_{v}, e_{t}])^{T}}{\sqrt{d_{k}}}),
\end{equation}
where $d_{k}$ is a scaling facter.
The attention matrix $A$ comprises elements $A_{ij}$ with $i, j \in {1, 2, \ldots, n}$, representing the relationship between the $i$-th token and the $j$-th token, and their impact on the output.
The exact visual encoder, LLM decoder, image-text alignment module and pretraining method for parameters $\phi$, $\theta$, and $\epsilon$ differ between models but the overarching method described above remains the same.

\subsection{Coarse-grained Location Grounding}

In this stage, we leverage the grounding capability of MLLM to locate the answer in the image.
As shown in Fig.~\ref{method}, given the image \textit{I} and the corresponding question \textit{Q}, we employ a manually designed grounding prompt \textit{P$_\text{g}$} (e.g. “According to the information in the image and the question, detail the bounding box of the region in the image that contains the answer in JSON format.”) to guide the MLLM to identify the approximate area in the image related to the question and output its corresponding bounding box coordinates.
Following the acquisition of the answer region's coordinates, we implement a post-processing step aimed at optimizing this output for the subsequent stage of attention reweighting, which generates the accurate response to question $Q$.
The post-processing process is described in the following.

The initial step involves determining the central coordinates of the bounding box and adjusting the bounding box based on them. 
This process introduces an expanding hyper-parameter $\alpha$, which dictates the appropriate size to which the bounding box should be expand to.
For cases that exceed the image boundary, we move it to keep it within the image. 

At this stage, we have acquired the coarse-grained position coordinates of the answer area, which contain sufficient information to generate the correct response and will be utilized to guide the reweighting of visual attention in the second stage.

\subsection{Fine-grained Attention Reweighting}

After obtaining the coordinates of the target region in the first stage, we convert it into a binary mask matrix $M$. 
Specifically, for elements within the target region, we assign a value of 1, while assigning 0 to all others.
This binary mask serves as a condition for decoding the subsequent sequence.
In practice, for each attention layer in the decoder, we multiply the attention scores of the image tokens corresponding to $M$ by a scaling factor $\lambda$, directing the model's focus toward a specific image region. 
The modified attention map $\hat{A}$ is formulated as:
\begin{equation}
\hat{A} = softmax(log(\lambda) \cdot M + A),
\label{eq:reweight}
\end{equation}
where $M$ is the binary mask matrix and $A$ denotes the origin attention score matrix.
The attention reweighting strategy activates the attention score of specific tokens in the attention modules, thereby effectively shift the focus of the model to specific tokens to achieve more detailed and precise control over the output.
Equation \ref{eq:reweight} presents the SoftMax probability on the activated fraction, with a consistent multiple increase in probability on specific visual tokens.
Through the attention reweighting mechanism, the model pays higher attention to the answer region to generate accurate answers to the question.

\begin{table*}[htbp]
\caption{Evaluations on comprehensive VLM benchmarks, including MME, MMBench and POPE.}
\centering
\setlength{\tabcolsep}{2.9mm}
\begin{tabular}{l c c c c c c c c c}
\toprule
\multirow{2.4}{*}{\textbf{Method}} &\multirow{2.4}{*}{\textbf{Projector}} &\multicolumn{2}{c}{\textbf{MME}} &\multicolumn{2}{c}{\textbf{MMBench}} &\multicolumn{3}{c}{\textbf{POPE}} &\multirow{2.4}{*}{\textbf{\textit{Sum}}}\\
\cmidrule(lr{0pt}){3-4} 
\cmidrule(lr{0pt}){5-6}
\cmidrule(lr{0pt}){7-9}
& &\textbf{\textit{Perception}} &\textbf{\textit{Cognition}} &\textbf{\textit{Dev}} &\textbf{\textit{Test}} &\textbf{\textit{Random}} &\textbf{\textit{Popular}} &\textbf{\textit{Adversarial}} \\
\midrule
BLIP2-13B~\cite{blip2} &Q-former &1293.8 &290.0 &- &- &89.6 &85.5 &80.9 &- \\
MiniGPT-4-7B~\cite{zhu2023minigpt} &Resampler &581.7 &144.3 &23.0 &- &- &- &- &- \\
Shikra-13B~\cite{chen2023shikra} &Linear &- &- &58.8 &-&- &- &- &- \\
Qwen-VL-chat-7B~\cite{bai2023qwen} &Resampler &1487.5 &360.8 &60.6 &- &- &- &- &- \\
mPLUG-Owl2-7B~\cite{ye2024mplug} &Resampler &1450.2 &- &64.5 &- &- &- &- &- \\
\midrule
LLaVA-v1.5-7B~\cite{liu2024improved} &MLP &1510.7 &285.0 &64.3 &66.4 &87.3 &86.1 &84.2 &2184.0 \\
\rowcolor{gray!10}
\textbf{LLaVA-v1.5-7B+CoF} &MLP &\textbf{1515.9} &\textbf{334.0} &\textbf{65.1} &\textbf{67.7} &\textbf{87.4} &\textbf{86.3} &\textbf{84.2} &\textbf{2240.6}\(_{\textcolor{blue}{+56.6}}\) \\
LLaVA-v1.5-13B~\cite{liu2024improved} &MLP &1531.3 &295.4 &67.7 &67.0 &87.1 &86.2 &84.5 &2219.2 \\
\rowcolor{gray!10}
\textbf{LLaVA-v1.5-13B+CoF} &MLP &\textbf{1545.6} &\textbf{310.7} &\textbf{68.3} &\textbf{69.2} &\textbf{88.0} &\textbf{86.6} &\textbf{84.8} &\textbf{2253.2}\(_{\textcolor{blue}{+34.0}}\) \\
InstructBLIP-13B~\cite{dai2023instructblip} &Q-former &1212.8 &\textbf{291.8} &29.2 &36.7 &\textbf{87.7} &77.0 &72.0 &1807.2 \\
\rowcolor{gray!10}
\textbf{InstructBLIP-13B+CoF} &Q-former &\textbf{1236.1} &290.8 &\textbf{29.2} &\textbf{50.6} &87.5 &\textbf{84.8} &\textbf{82.5} &\textbf{1861.5}\(_{\textcolor{blue}{+54.3}}\) \\
\bottomrule
\end{tabular}
\label{main}
\end{table*}

\begin{table}[htbp]
\caption{Ablation studies on LLaVA-v1.5-13B model.}
\centering
\setlength{\tabcolsep}{2.60mm}
\begin{tabular}{c c c c c c}
\toprule
\multirow{2.4}{*}{\textbf{Reweight}} &\multirow{2.4}{*}{\textbf{Ground}} &\multicolumn{2}{c}{\textbf{MME}} 
&\multicolumn{2}{c}{\textbf{MMBench}} \\
\cmidrule(lr{0pt}){3-4}
\cmidrule(lr{0pt}){5-6} 
& &\textbf{\textit{Perception}} &\textbf{\textit{Cognition}} &\textbf{\textit{Dev}} &\textbf{\textit{Test}}\\
\midrule
& &1531.3 &295.4 &67.7 &67.0 \\
\checkmark & &1527.2 &306.8 &\textbf{68.5} &69.1 \\
\checkmark &\checkmark &\textbf{1545.6} &\textbf{310.7} &68.3 &\textbf{69.2} \\
\bottomrule
\end{tabular}
\label{ablation}
\end{table}

\section{Experiment}

\subsection{Experimental Setup}
We apply our CoF approach to the popular LLaVA-v1.5-7B, LLaVA-v1.5-13B \cite{liu2024improved} and InstructBLIP-13B \cite{dai2023instructblip}.
In the query-based mapping method InstructBLIP-13B, we apply the attention reweighting method to Q-former.
These models are evaluated on benchmarks (i.e., MME~\cite{MME}, MMBench~\cite{mmbench}, POPE~\cite{pope}), focusing on multi-modal understanding and hallucination tasks.
MME measures both perception and cognition abilities on a total of 14 subtasks.
MMBench is a systematically designed objective benchmark for a robust and holistic evaluation of MLLM.
POPE is a polling-based query method for better evaluation of object hallucination.
Regarding hyperparameter settings, the scaling hyperparameter $\alpha$ is set to 1.3, and $\lambda$ to 2.0 for the LLaVA-v1.5-7B model. 
For the LLaVA-v1.5-13B model, $\alpha$ is set to 1.0 while $\lambda$ is adjusted to 4.5. 
In the case of InstructBLIP-13B, $\alpha$ is set to 1.0 and $\lambda$ is set to 22.0. 
We performed all experiments on an NVIDIA-A800-80GB and a Tesla-V100-sxm2-32GB GPU.

\subsection{Experimental Results}

Our primary experimental results are summarized in Table~\ref{main}. 
The data indicate a marked improvement in the performance of our approach across three benchmarks, demonstrating its effectiveness.
Specifically, the results on the MME and MMBench benchmarks highlight advancements in multi-modal understanding tasks. 
Furthermore, the enhancement observed in the POPE benchmark underscores the success of CoF in mitigating hallucinations in MLLM.
This supports our hypothesis that enhancing the model's grounding capabilities contributes to improved fine-grained understanding and reduces the effects of hallucinations.
Significant advancements were achieved in the MME perception task, which illustrates the superiority of our coarse to fine-grained perception approach over traditional fixed-grained methods. 
These results illustrate the model’s ability to accurately identify the relevant answer areas, thereby streamlining the inference process from the original image and enhancing the overall performance of the model.

\subsection{Ablation Study}

In this section, we present comprehensive ablation experiments on the LLaVA-v1.5-13B model. 
The results are summarized in Table~\ref{ablation}.
We first evaluate the efficacy of the attention reweighting mechanism by scaling the attention scores of all image tokens. 
Previous research indicates that as the length of the output sequence increases, the model's reliance on visual prompts diminishes, resulting in a tendency to “forget” input image information~\cite{favero2024multi}. 
Consequently, the model may over-rely on previously generated text sequences to predict answers, making it more susceptible to hallucinations due to text priors~\cite{favero2024multi}. 
Compared with the baseline, the introduction of the attention reweighting mechanism emphasizes the importance of visual prompts in answer prediction,  guiding the model to predict answers that are more faithful to the image, thereby improving performance on the benchmarks. 
However, this variant still processes image information at a fixed granularity and lacks local fine-grained visual attention guidance, making it easily affected by background noise.

To address this limitation, we introduced the location grounding stage to direct the model's focus toward the relevant local context. 
Previous studies employed the grounding capability of MLLM to define the answer area by cropping and scaling it~\cite{shao2024visual, luan2024textcot}, which often compromises the overall semantic integrity of the image. 
This is particularly detrimental when the model exhibits relatively weak grounding capabilities, as it can lead the model to predict incorrect answers.
However, CoF successfully preserves the global semantic information of the image while simultaneously highlighting the targeted local area, thereby diminishing the occurrence of false positive predictions. 
Ultimately, CoF operates through the aforementioned two-stage approach, transitioning from coarse-grained to fine-grained image perception while retaining comprehensive semantic information, which contributes to achieving optimal performance.

\section{Conclusion}

In this paper, we proposed a step-by-step multi-modal understanding approach from coarse to fine (CoF). 
By decomposing the complex image understanding task into two stages: coarse-grained localization and fine-grained recognition, our CoF approach improved the overall performance on multiple downstream tasks, verifying its feasibility. 
In addition, the imitation of the human visual perception process makes CoF more interpretable. 
In the future, we will continue to study the mechanism of attention in the process of image and text interaction, and move forward towards improving the fine-grained understanding ability of the model and continue to explore the interpretability of MLLM.

\clearpage

\bibliographystyle{IEEEtran}
\bibliography{IEEEexample}

\end{document}